%% file: main.tex
\documentclass[sigconf]{acmart}

\setcopyright{none}
\renewcommand\footnotetextcopyrightpermission[1]{}

\usepackage{enumitem}
\usepackage{xcolor}
\usepackage{tcolorbox}

\copyrightyear{2026}
\acmYear{2026}
\setcopyright{usgov}
\acmConference[DocEng '26]{Proceedings of the 2026 ACM Symposium on Document Engineering}{August 25--28, 2026}{Fribourg, Switzerland}
\acmBooktitle{Proceedings of the 2026 ACM Symposium on Document Engineering (DocEng '26), August 25--28, 2026, Fribourg, Switzerland}
\acmDOI{10.1145/3820755.3821487}
\acmISBN{979-8-4007-2786-3/2026/08}

\begin{document}
\title[Prompt Programming for Cultural Bias and Alignment of LLMs]{Prompt Programming for \\Cultural Bias and Alignment of Large Language Models}

\author{Maksim E. Eren}
\authornote{Both authors contributed equally to this work.}
\email{maksim@lanl.gov}
\orcid{0000-0002-4362-0256}
\affiliation{%
    \institution{Computational Intelligence \& Modeling, Los Alamos National Laboratory}
    \city{Los Alamos}
    \state{New Mexico}
    \country{USA}
}

\author{Eric Michalak}
\authornotemark[1]
\email{emichalak@lanl.gov}
\affiliation{%
    \institution{Advanced Research in Cyber Systems, Los Alamos National Laboratory}
    \city{Los Alamos}
    \state{New Mexico}
    \country{USA}
}

\author{Brian Cook}
\email{cook_b@lanl.gov}
\affiliation{%
    \institution{Center for National Security and International Studies}
    \city{Los Alamos}
    \state{New Mexico}
    \country{USA}
}

\author{Johnny Seales Jr.}
\email{jseales@lanl.gov}
\affiliation{%
    \institution{Analytics Division, Los Alamos National Laboratory}
  \city{Los Alamos}
  \state{New Mexico}
  \country{USA}
}

\renewcommand{\shortauthors}{Eren and Michalak et al.}
\begin{abstract}
Culture shapes reasoning, values, prioritization, and strategic decision-making, yet large language models (LLMs) often exhibit cultural biases that misalign with target populations. As LLMs are increasingly used for strategic decision-making, policy support, and document engineering tasks such as summarization, categorization, and compliance-oriented auditing, improving cultural alignment is important for ensuring that downstream analyses and recommendations reflect target-population value profiles rather than default model priors. Previous work introduced a survey-grounded cultural alignment framework and showed that culture-specific prompting can reduce misalignment, but it primarily evaluated proprietary models and relied on manual prompt engineering. In this paper, we \textit{validate and extend} that framework by reproducing its social sciences survey based projection and distance metrics on open-weight LLMs, testing whether the same cultural skew and benefits of culture conditioning persist outside closed LLM systems. We then introduce use of prompt programming with \emph{DSPy} for this problem—treating prompts as modular, optimizable programs—to tune cultural conditioning by optimizing against cultural-distance objectives. In our experiments, we show that prompt optimization often improves upon cultural prompt engineering, suggesting prompt compilation with \emph{DSPy} can provide a more stable and transferable route to culturally aligned LLM responses.

\end{abstract}

\ccsdesc[500]{Computing methodologies~Artificial intelligence}
\ccsdesc[500]{Computing methodologies~Natural language processing}
\ccsdesc[300]{Computing methodologies~Machine learning}

\keywords{LLM, Culture, Bias, Prompt Engineering, Prompt Programming}

\maketitle


\section{Introduction}
\input{sections/intro}

\section{Related Work}
\input{sections/related_work}

\section{Methods}

\input{sections/methods}

\section{Results}
\input{sections/results}

\section{Discussion and Future Work}
\input{sections/discussions}

\section{Conclusion}
\input{sections/conclusion}

\begin{acks}
This manuscript has been approved for unlimited release and has been assigned LA-UR-26-21996. The funding for this paper was provided by Los Alamos National Laboratory (LANL). LANL is operated by Triad National Security, LLC, for the National Nuclear Security Administration of the U.S. Department of Energy (Contract No. 89233218CNA000001).
\end{acks}
\bibliographystyle{ACM-Reference-Format}
\bibliography{sections/ref}
\end{document}

%% file: sections/intro.tex
Cultural systems shape how individuals interpret uncertainty, prioritize social goals, evaluate authority, and resolve moral trade-offs. These dimensions influence strategic decision-making in domains such as governance, organizational leadership, education, and public communication. For the purpose of this paper, we define culture as a system of societal norms and the way of life that are learned and shared among major groups in a population~\cite{Hofstede2001CulturesCC, pgae346}. As large language models (LLMs) increasingly assist in drafting policies, generating recommendations, categorizing and analyzing documents, and mediating cross-cultural interactions, their implicit value orientations can meaningfully affect downstream reasoning. Culture can also shape document engineering itself: it affects how prompts, templates, schemas, and response constraints define salient categories, acceptable evidence, and legitimate justifications. Because LLMs are increasingly used for document engineering tasks such as summarization, categorization, classification, and auditing support, these prompt- and template-based artifacts function as interaction specifications through which cultural assumptions enter into what a system requests, prioritizes, and treats as a valid rationale. Cultural misalignment in model outputs is therefore not merely representational; it can shift which options are surfaced, how trade-offs are justified, and what is treated as legitimate or desirable in downstream document workflows.

Tao et al.~\cite{pgae346} introduced a social-science-based, survey-grounded framework for measuring cultural alignment by mapping LLM responses to standardized value instruments and computing distance from nationally representative benchmarks. They showed that, under generic prompting, models concentrate around Western value profiles, and that explicit culture conditioning through prompt engineering can reduce misalignment by moving LLM responses toward target countries/territories. Although influential, Tao et al.~\cite{pgae346} evaluated only proprietary models and relied solely on manual prompt engineering. In contrast, open-weight models offer research advantages that closed systems typically cannot---democratized access, reproducibility, and broader support for diverse downstream applications~\cite{Manchanda2024TheOS}. Moreover, while prompt engineering can improve LLM outputs~\cite{Schulhoff2024ThePR}, it is increasingly complemented, and in some settings supplanted, by programmatic prompt optimization frameworks such as \emph{DSPy}~\cite{khattab2023dspy}, which treat prompts as optimizable components and automatically search for prompts that best satisfy an explicit objective and format constraints. In this sense, prompt optimization is relevant to document engineering because prompts and response constraints act as specifications for LLM-based document workflows, shaping how generated documents are structured, what information is emphasized, and which justifications are treated as acceptable. Our work here, therefore, addresses two research questions:
\begin{enumerate}
\item Do the findings of Tao et al.~\cite{pgae346} hold for open-weight LLMs?
\item Does prompt optimization culturally align LLM responses better than prompt engineering?
\end{enumerate}

To address the first research question, we reproduce the survey-grounded cultural-alignment pipeline introduced by Tao et al.~\cite{pgae346} on open-weight models spanning scales and training regimes: Llama~3.3 (70B)~\cite{llama3_2024}, Llama~4 (16$\times$17B)~\cite{llama4_2024}, Gemma~3 (27B)~\cite{gemma3_2024}, and GPT-OSS (20B/120B)~\cite{gptoss_2024}. Across these models, we observe that culture conditioning remains detectable and consistent, indicating that Tao et al.'s findings extend beyond proprietary systems and persist across architectures and scales. To address the second research question, we compare manual cultural prompt engineering against \emph{DSPy}-based prompt optimization, which treats prompting as a parameterized program and compiles prompts against explicit objectives under strict response-format constraints~\cite{khattab2023dspy}. By casting cultural alignment as an optimization problem---minimizing human survey-derived cultural distance---we find that \emph{DSPy} moves responses closer to target cultural benchmarks than manual prompting in most, but not all, evaluated settings. Together, these results suggest that programmatic prompt optimization can provide a more systematic route to cultural conditioning across open-weight models, while also motivating downstream validation in culturally sensitive document-engineering tasks such as summarization, categorization, and auditing. In summary, our contributions include:
\begin{enumerate}
\item Validating and extending the prior work by Tao et al.~\cite{pgae346} on five different open-weight LLMs.
\item Introducing use of prompt programming with \emph{DSPy}~\cite{khattab2023dspy} for cultural alignment of LLMs and comparing its performance to prompt engineering.
\item Evaluating the cultural alignment performance of prompt programming using two \emph{DSPy} teleprompters, Cooperative Prompt Optimization (COPRO) and Multiprompt Instruction Proposal Optimizer, version 2 (MIPROv2)~\cite{dspy_copro_docs, dspy_miprov2_docs}, and comparing the impact of a small vs.\ large instruction-suggester model during optimization (Llama~3.2 1B vs.\ GPT-OSS~120B).
\end{enumerate}

%% file: sections/related_work.tex
Tao et al.~\cite{pgae346} introduced a survey-grounded framework for measuring cultural alignment by mapping LLM responses to World Values Survey (WVS) items into the Inglehart--Welzel (IW) cultural map and computing distance to nationally representative benchmarks. Their evaluation of proprietary OpenAI models showed that generic prompting induces a consistent Western concentration in responses, while culture-specific prompting can reduce cultural distance for many targets. Our work builds on this framework by replicating the full pipeline on open-weight models and introducing \emph{DSPy} prompt programming as an alternative to manual cultural prompt engineering.

As LLMs are incorporated into document-centric systems, \emph{document engineering}---spanning multi-document summarization, categorization pipelines, auditing workflows, and the construction of prompts, templates, schemas, and documentation artifacts~\cite{Godbole2024LeveragingLL,Yao2024SmartAS}---provides the interface through which models are applied. These artifacts are not culturally invariant: schema and template choices shape which distinctions are encoded and what counts as evidence, and summarization can introduce systematic distortions in what is selected and emphasized~\cite{Steen2023BiasIN}. More broadly, task and evaluation design embed cultural assumptions~\cite{Oh2025CultureIE}, and prompt language or explicit cultural framing shifts expressed values, indicating that interaction specifications can act as cultural cues~\cite{Bulte2025LLMsAC}. Further, documentation supports auditability and certification, with scope varying by perceived risk~\cite{KONIGSTORFER2022100043}. From a cross-cultural tooling perspective, O'Neil et al.~\cite{oneil-etal-2024-computational} emphasize linguistic sovereignty, cultural specificity, and reciprocity, while Bravansky et al.~\cite{Bravansky2025RethinkingAC} frame cultural alignment as context-dependent and shaped by interaction scaffolding. Taken together, these findings imply that as LLMs are increasingly used for document analysis, their cultural priors and biases can shape summarization, categorization, and auditing behaviors within document-engineering workflows.

Subsequent work has expanded survey-based evaluation and persona conditioning at scale. Zhao et al.~\cite{Zhao2024WorldValuesBenchAL} introduce \emph{WorldValuesBench} (WVB), a large-scale benchmark derived from WVS Wave~7 with over 20 million $(\text{demographics}, \text{question}) \rightarrow \text{answer}$ instances, and evaluate models using Wasserstein distance to human answer distributions. AlKhamissi et al.~\cite{AlKhamissi2024InvestigatingCA} simulate sociological surveys across persona specifications and show that cultural misalignment varies systematically with prompt language and demographic conditioning, proposing anthropological prompting as a mitigation strategy. Tuna et al.~\cite{10852463} similarly compare GPT-3.5-turbo and GPT-4 responses across five language areas and ten subcultures, finding substantial variation in cultural alignment by linguistic region and subculture specification. Kwok et al.~\cite{kwok2024evaluating} evaluate synthetic personas against real participants and show that nationality conditioning improves alignment, while native-language prompting alone does not reliably do so. Complementing these, Greco et al.~\cite{Greco2026CulturallyGP} generate WVS-grounded synthetic personas and analyze their placement on the IW map and Moral Foundations Theory (MFT) dimensions, demonstrating structured cross-cultural variation but also highlighting representation biases in persona generation.

Broader analyses of cultural skew contextualize these findings. Atari et al.~\cite{atari2023which} query OpenAI models with WVS items models and show that responses cluster closest to Western, Educated, Industrialized, Rich, and Democratic (WEIRD) societies, with similarity decreasing as countries become more culturally distant from the United States. Zhou et al.~\cite{Zhou2025ShouldLB} further examine this \emph{WEIRDness}--rights trade-off, using the Universal Declaration of Human Rights (UDHR) and regional charters to show that models less aligned with WEIRD populations produce more culturally variable responses but are more likely to generate outputs that conflict with human-rights principles. Arora et al.~\cite{arora2023probing} evaluate multilingual pretrained language models (PLMs) using WVS- and Hofstede-derived templates and find that while models encode cross-cultural variation, their induced value rankings correlate weakly with survey ground truth and vary across architectures. At a systems level, Pawar et al.~\cite{Pawar2024SurveyOC} survey work on cultural awareness in language and multimodal models, emphasizing that evaluation remains fragmented across datasets, modalities, and alignment objectives, motivating more consistent cross-cultural benchmarking grounded in social-scientific theory.

Complementary task-based benchmarks document operational considerations of cultural perspective. Naous et al.~\cite{naous2024beer} introduce the Cultural Awareness in Arabic language Models benchmark (\emph{CAMeL}) and show that multilingual models systematically prefer Western-associated entities across generation, infilling, named entity recognition (NER), and sentiment analysis, with prompt adaptations reducing but not eliminating these effects. Navigli et al.~\cite{navigli2023biases} argue that such disparities often originate in upstream data selection bias, including domain, temporal, and demographic skew, and advocate data-centric interventions alongside model-level mitigation. Similarly, Johnson et al.~\cite{johnson2022ghost} examine GPT-3 through summarization tasks and document shifts toward mainstream United States positions, consistent with training-data dominance effects~\cite{naous2024beer}. Together with distributional distance analyses, these results suggest that cultural skew can influence not only surface-level outputs but also which trade-offs are framed as legitimate or desirable.

Work in psycholinguistic modeling reinforces that culturally inflected language patterns correlate with consequential institutional outcomes. Sigdel and Panfilova~\cite{sigdel2026ruslica} develop \emph{RusLICA}, a Russian-language adaptation of Linguistic Inquiry and Word Count (LIWC) incorporating morphology-aware lexical resources, highlighting that culturally aligned operationalization of linguistic categories is necessary for valid inference. In an applied governance setting, Gandall et al.~\cite{Gandall2022PredictingPA} show that LIWC-derived rhetorical and psychological features in United Nations (UN) deliberations predict policy passage and coercive-intervention context substantially better than procedural-motion baselines. These findings imply that  linguistic signals correlate with institutional decisions, motivating careful measurement of cultural alignment when LLMs are used in policy-adjacent contexts.

Finally, strategic and agentic evaluations demonstrate that LLMs increasingly participate in decision-making environments where value orientation can shape outcomes. Bakhtin et al.~\cite{Bakhtin2022HumanlevelPI} introduce \emph{Cicero}, an agent that achieves human-level performance in the negotiation game \emph{Diplomacy} by coupling intent-conditioned dialogue generation with explicit strategic planning over belief states and joint actions. Payne~\cite{payne2026ai} extends this line to nuclear crisis simulations, showing that frontier models exhibit structured strategic reasoning, including deception, theory-of-mind modeling, and stable ``strategic personalities,'' with escalation dynamics sensitive to temporal framing. Similarly, Hogan and Brennen \cite{Hogan2024OpenEndedWW} introduce \emph{Snow Globe}, an open-source framework for simulating and playing open-ended wargames with multi-agent LLMs for understanding real-world decision making. As LLMs move from passive text generators to agents embedded in planning loops, governance, and scientific workflows, their implicit value priors can influence which options are surfaced, how commitments are framed, and how risk is evaluated. Gridach et al.~\cite{Gridach2025AgenticAF} survey agentic systems in scientific discovery and emphasize reliability, calibration, and ethical oversight as core challenges, while Eren and Perez~\cite{eren2025rethinking} argue that artificial intelligence (AI) systems increasingly shape hypothesis generation, literature synthesis, and collaborative decision-making under information overload. In such settings, cultural misalignment is not merely representational: it can structure reasoning trajectories and affect policy, governance, and scientific judgments, motivating our work on building up on and extending the work from Tao et al.~\cite{pgae346} with prompt optimization for cultural alignment, and testing it with across open-weight models.

%% file: sections/methods.tex
We largely follow the IVS-based cultural alignment pipeline of Tao et al.~\cite{pgae346}, replicating their benchmark projection and distance metrics, and then extending the evaluation to open-weight models and \emph{DSPy}-based prompt programming. This section first describes the replicated baseline pipeline in full for completeness, then our \emph{DSPy}-based prompt-programming additions are detailed in Section ~\ref{sec:dspy_methods}.

\subsection{Cultural map of countries/territories}
\label{sec:cultural_map}

We first replicate the cultural map of countries/territories as described by Tao et al. \cite{pgae346}, which we summarize here for completeness. First, we construct the Integrated Values Surveys (IVS) benchmark cultural space and the country/territory reference locations using the IVS, a harmonized integration of World Values Survey (WVS) and European Values Study (EVS) data \cite{ivs2023,haerpfer2022wvs,evs2022trend}. Following the survey-grounded cultural alignment framework of Tao et al. \cite{pgae346}, we focus on the three most recent joint waves (2005--2022) and retain all available country--wave observations within this window. We use the same ten survey indicators from prior work, that build the Inglehart--Welzel cultural map, including but not limited to survey questions on happiness, social trust, authority, petition signing, national pride, and autonomy \cite{inglehartwelzel2005,pgae346}. Question on happiness as prompt example is included in Box \ref{box:example_prompt}. Responses are transformed into numeric variables using the IVS/WVS/EVS coding guidance \cite{ivs2023,haerpfer2022wvs,evs2022trend,pgae346}.

We fit Principal Component Analysis (PCA) on standardized respondent-level values across the ten indicators, apply varimax rotation, and incorporate survey weights by using survey-weighted moments for standardization and survey-weighted aggregation for country--wave means \cite{jolliffe2016pca,kaiser1958varimax,pgae346}. The first two rotated components are interpreted as the canonical axes \emph{Survival vs.\ Self-Expression} and \emph{Traditional vs.\ Secular} \cite{inglehartwelzel2005,pgae346}. To match the benchmark coordinate system, we apply the same linear rescaling used by Tao et al. \cite{pgae346}:
\begin{align}
\text{PC1}' &= 1.81 \cdot \text{PC1} + 0.38, \\
\text{PC2}' &= 1.61 \cdot \text{PC2} - 0.01.
\end{align}
Country/territory reference points $\boldsymbol{\nu}^{\text{IVS}}_{c} \in \mathbb{R}^2$ are computed by first taking survey-weighted means within each country--wave and then averaging the resulting country--wave points across waves in 2005--2022.\footnote{We average across waves with equal weight per country--wave observation, consistent with treating each observed country--wave as one estimate of the country’s position in the shared IVS space.} We visualize these country/territory reference points as the IVS cultural map (Figure~\ref{fig:cultural_map_opensource}).

\subsection{Open-weight model projection into the IVS benchmark space}
\label{sec:model_projection}

With the benchmark space established, we next project open-weight LLM responses into the same coordinate system to enable direct comparison to the IVS country/territory map, replicating the approach of Tao et al. \cite{pgae346}, but on open-weight models rather than the closed-source systems emphasized there. We evaluate five open-weight LLMs: Llama~3.3 (70B) \cite{llama3_2024}, Llama~4 (16x17B) \cite{llama4_2024}, Gemma~3 (27B) \cite{gemma3_2024}, GPT-OSS (20B), and GPT-OSS (120B) \cite{gptoss_2024}. For each of the ten IVS indicators, we prompt models with \textbf{(i)} a short persona statement and \textbf{(ii)} the survey question text paired with response-format constraints aligned with the survey coding scheme (e.g., one integer or one categorical option), enabling deterministic mapping from model outputs to the corresponding numeric variables \cite{pgae346}. In the \emph{generic} setting (no cultural prompting), we ask the IVS survey question directly (with the same response-format constraints). In the \emph{cultural} setting, we prepend the survey question with a lightweight country/territory prefix of the form \textcolor{blue}{``You are a citizen of \texttt{X}.''} to condition responses on the target cultural context.

\begin{tcolorbox}[title={Example prompting (generic vs.\ cultural). \\
A008 (Feeling of Happiness) Question from IVS}, colback=white, colframe=black]
\small
\textbf{Generic (no cultural prompting):}\\
\texttt{Question: Taking all things together, rate how happy you would say you are.}\\
\texttt{Please use a scale from 1 to 4, where 1 is Very happy, 2 is Quite happy,}\\
\texttt{3 is Not very happy, 4 is Not at all happy.}\\
\texttt{You can only respond with a score number based on the scale provided and please do not give reasons.}\\
\texttt{Your score number:}

\textbf{\textcolor{blue}{Cultural prompting (country/territory-conditioned):}}\\
\textcolor{red}{\texttt{You are a citizen of United States of America (USA).}}\\
\textcolor{blue}{\texttt{Question: Taking all things together, rate how happy you would say you are.}}\\
\textcolor{blue}{\texttt{Please use a scale from 1 to 4, where 1 is Very happy, 2 is Quite happy,}}\\
\textcolor{blue}{\texttt{3 is Not very happy, 4 is Not at all happy.}}\\
\textcolor{blue}{\texttt{You can only respond with a score number based on the scale provided and please do not give reasons.}}\\
\textcolor{blue}{\texttt{Your score number:}}
\label{box:example_prompt}
\end{tcolorbox}

Again, following Tao et al. \cite{pgae346}, we explicitly account for prompt-phrasing variance by evaluating a small set of semantically equivalent respondent-descriptor personas and averaging results to reduce sensitivity to wording choices; concretely, we vary the respondent descriptor using synonymous terms such as \textcolor{blue}{\texttt{average}/\texttt{typical}}, \textcolor{blue}{\texttt{human being}/\texttt{person}/\texttt{individual}}, and \textcolor{blue}{\texttt{world citizen}} and average the resulting projected coordinates across variants. We use deterministic decoding (temperature $=0$) and minimal output sanitation to extract the required option \cite{pgae346}.

\begin{figure*}[htb]
  \centering
  \includegraphics[width=0.9\textwidth]{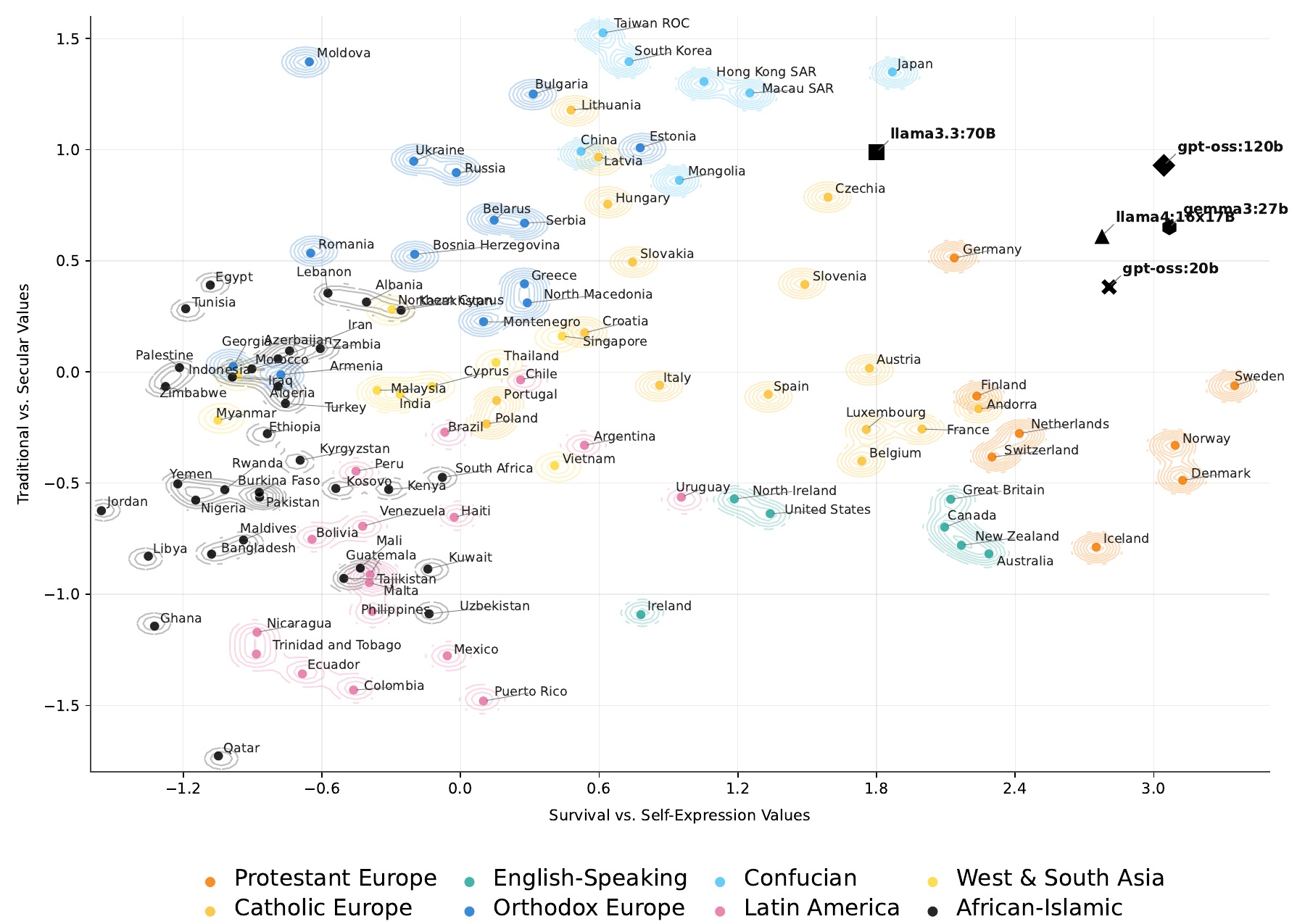}
  \caption{Cultural map of countries/territories in the IVS benchmark space (Survival vs.\ Self-Expression; Traditional vs.\ Secular values). We overlay points derived from open-weight model responses (Llama~3.3 70B, Llama~4 16$\times$17B, Gemma~3 27B, GPT-OSS 20B, GPT-OSS 120B) to the same IVS items, using the projection procedure introduced by Tao et al.~\cite{pgae346}. Shaded regions are visual guides highlighting approximate cultural-zone clusters and are not used in distance calculations.}
  \label{fig:cultural_map_opensource}
\end{figure*}

Let $\mathbf{x}_{m,c,v} \in \mathbb{R}^{10}$ be the coded response vector produced by model $m$ under country/territory condition $c$ and persona variant $v$, with $c=\varnothing$ denoting a non-national generic persona. We project each response vector into the IVS benchmark space by standardizing with IVS-derived moments and applying the IVS-fitted \emph{rotated} two-dimensional scoring map:
\begin{align}
\mathbf{z}_{m,c,v}
&= \left(\mathbf{x}_{m,c,v} - \boldsymbol{\mu}^{\text{IVS}}_{\text{raw}}\right)
   \oslash \boldsymbol{\sigma}^{\text{IVS}}_{\text{raw}}, \\
\mathbf{s}_{m,c,v}
&= \mathbf{W}_{\text{rot}}\,\mathbf{z}_{m,c,v},
\end{align}
where $\boldsymbol{\mu}^{\text{IVS}}_{\text{raw}} \in \mathbb{R}^{10}$ and $\boldsymbol{\sigma}^{\text{IVS}}_{\text{raw}} \in \mathbb{R}^{10}$ are the (survey-weighted) IVS means and standard deviations for the ten indicators, $\oslash$ denotes elementwise division, and $\mathbf{W}_{\text{rot}}\in\mathbb{R}^{2\times 10}$ are the first-two-component \emph{varimax-rotated} PCA scoring weights estimated from IVS data \cite{pgae346}. We then rescale $\mathbf{s}_{m,c,v}$ using Eqs.~(1--2) to obtain $\boldsymbol{\pi}_{m,c,v} \in \mathbb{R}^{2}$ and average across persona variants:
\begin{align}
\boldsymbol{\mu}_{m,c} &= \frac{1}{|V|}\sum_{v\in V} \boldsymbol{\pi}_{m,c,v}.
\end{align}
We overlay $\boldsymbol{\mu}_{m,\varnothing}$ (generic, non-national prompting) for each open-weight model on the IVS cultural map to visualize how open-weight models cluster relative to country/territory reference locations (Figure \ref{fig:cultural_map_opensource}).

\subsection{Country-level cultural distance under three prompting regimes}
\label{sec:distance_regimes}

Having embedded both IVS respondents and model responses in a common space, we quantify alignment using country-level Euclidean distance in the IVS benchmark coordinates as described by Tao et al. \cite{pgae346}. Let $\boldsymbol{\nu}^{\text{IVS}}_{c}\in\mathbb{R}^{2}$ denote the IVS reference coordinate for country/territory $c$ (computed from human respondents). All distances are measured to the same human reference point $\boldsymbol{\nu}^{\text{IVS}}_{c}$.

We compare three prompting regimes for cultural conditioning:

\textbf{(1) No culture conditioning.} We compute a single generic model point $\boldsymbol{\mu}_{m,\varnothing}$ and compare it to every country benchmark:
\begin{align}
d_{\text{gen}}(m,c)
&= \left\lVert \boldsymbol{\mu}_{m,\varnothing} - \boldsymbol{\nu}^{\text{IVS}}_{c} \right\rVert_2.
\label{eq:nocond_distance}
\end{align}

\textbf{(2) Manual culture prompting.} We condition the model on country/territory identity $c$ via a fixed cultural prefix while keeping the survey questions and answer constraints unchanged \cite{pgae346}. This yields $\boldsymbol{\mu}^{\text{man}}_{m,c}$ and distance
\begin{align}
d_{\text{man}}(m,c)
&= \left\lVert \boldsymbol{\mu}^{\text{man}}_{m,c} - \boldsymbol{\nu}^{\text{IVS}}_{c} \right\rVert_2.
\label{eq:manual_distance}
\end{align}

\textbf{(3) Culture prompt programming (\emph{DSPy}).} We replace the fixed manual prefix with a compiled prompt program produced by DSPy \cite{khattab2023dspy}, as described in Section \ref{sec:dspy_methods}, yielding $\boldsymbol{\mu}^{\text{DSPy}}_{m,c}$ and distance
\begin{align}
d_{\text{DSPy}}(m,c)
&= \left\lVert \boldsymbol{\mu}^{\text{DSPy}}_{m,c} - \boldsymbol{\nu}^{\text{IVS}}_{c} \right\rVert_2.
\label{eq:dspy_distance}
\end{align}

To summarize changes relative to the no-conditioning baseline, we report paired country-level distance reductions
\begin{align}
\Delta d_{\text{man}}(m,c)
&= d_{\text{gen}}(m,c) - d_{\text{man}}(m,c), \\
\Delta d_{\text{DSPy}}(m,c)
&= d_{\text{gen}}(m,c) - d_{\text{DSPy}}(m,c),
\end{align}
where positive values indicate that the prompted condition moves the model closer to the country/territory benchmark. We visualize the resulting country-level distance distributions in Figure~\ref{fig:cultural_distance_dspy}, with DSPy prompt-programming results reported on held-out test countries as described in Section~\ref{sec:dspy_methods}.

\subsection{Culture prompt programming with DSPy}
\label{sec:dspy_methods}

We use \emph{DSPy} for prompt programming because it provides a way to treat cultural conditioning as an optimization problem: instead of committing to a single hand-written persona template, we allow the culture-conditioning instruction to be automatically tuned to reduce country-level cultural distance \cite{khattab2023dspy}. Concretely, we parameterize the culture-conditioning instruction by a discrete prompt parameter $\theta$ (a text instruction) that is inserted upstream of the fixed IVS question block; for a given $(m,c,\theta)$, the resulting program elicits the ten IVS responses under the same response-format constraints used throughout and projects them into the IVS space to obtain $\boldsymbol{\mu}_{m,c}(\theta)\in\mathbb{R}^2$ (Sections~\ref{sec:model_projection}--\ref{sec:distance_regimes}) \cite{pgae346}. We train the compiler to reduce misalignment by directly optimizing country-level cultural distance to IVS benchmarks. Let $\boldsymbol{\nu}^{\text{IVS}}_{c}\in\mathbb{R}^{2}$ be the IVS benchmark coordinate for country/territory $c$, and let $\boldsymbol{\mu}_{m,c}(\theta)\in\mathbb{R}^{2}$ denote the persona-variant-averaged coordinate produced by the target model $m$ under the \emph{DSPy}-compiled prompt instruction $\theta$. We define the per-country objective
\begin{align}
d(m,c;\theta)
&= \left\lVert \boldsymbol{\mu}_{m,c}(\theta) - \boldsymbol{\nu}^{\text{IVS}}_{c} \right\rVert_2, \\
\text{score}(m,c;\theta)
&= -d(m,c;\theta),
\end{align}
so that maximizing $\text{score}$ is equivalent to minimizing cultural distance. In summary, our objective in tuning the prompt is to minimize the cultural distance defined in Section~\ref{sec:distance_regimes}. \emph{DSPy} compilation selects $\theta$ by maximizing the negative mean distance, equivalently minimizing mean cultural distance, over a training set of countries/territories $\mathcal{C}_{\text{train}}$:
\begin{align}
J(\theta; m, \mathcal{C}_{\text{train}})
&= \frac{1}{|\mathcal{C}_{\text{train}}|}
   \sum_{c\in \mathcal{C}_{\text{train}}} \text{score}(m,c;\theta) \notag\\
&= -\frac{1}{|\mathcal{C}_{\text{train}}|}
   \sum_{c\in \mathcal{C}_{\text{train}}} d(m,c;\theta), \\
\theta^\star
&= \arg\max_{\theta}\; J(\theta; m, \mathcal{C}_{\text{train}}) \notag\\
&= \arg\min_{\theta}\;
   \frac{1}{|\mathcal{C}_{\text{train}}|}
   \sum_{c\in \mathcal{C}_{\text{train}}} d(m,c;\theta).
\end{align}

\begin{figure*}[htb]
  \centering
  \includegraphics[width=\textwidth]{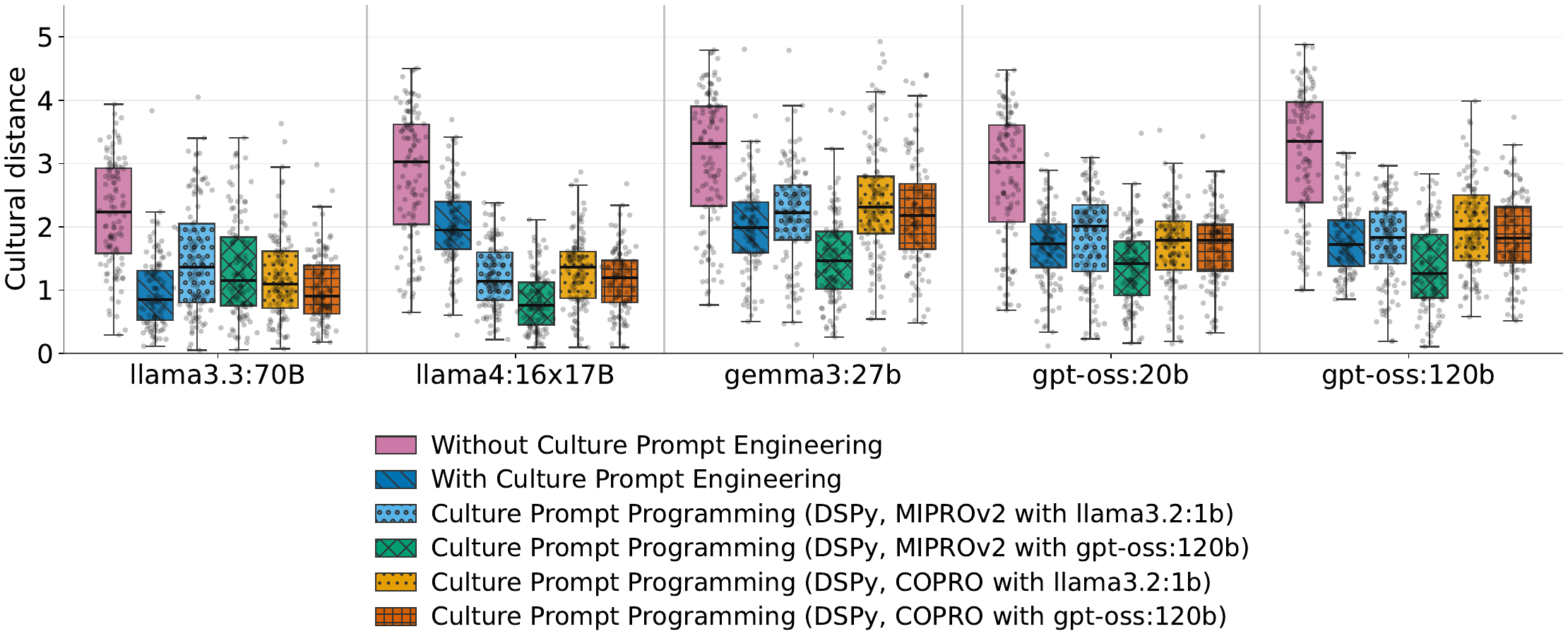}
  \caption{Country-level cultural distance for open-weight LLMs under three prompting regimes: (i) without culture conditioning, (ii) with manual culture prompt engineering, and (iii) culture prompt programming with DSPy. Cultural distance is measured, for each country/territory $c$, as the Euclidean distance between the model's projected coordinate in the IVS cultural-map space and the corresponding human reference coordinate $\boldsymbol{\nu}^{\mathrm{IVS}}_c$. For DSPy prompt-programming conditions, distances are computed on held-out test-set countries across 5-fold country-level cross-validation. The no-culture-prompt and manual culture-prompt conditions are baseline country-level distance distributions shown for comparison. Smaller distances indicate closer alignment to the country/territory benchmark, and changes across regimes reflect how cultural conditioning and prompt programming shift the model toward $\boldsymbol{\nu}^{\mathrm{IVS}}_c$.}
  \label{fig:cultural_distance_dspy}
\end{figure*}

We evaluate two \emph{DSPy} teleprompters, COPRO and MIPROv2 \cite{dspy_copro_docs,dspy_miprov2_docs}. Both methods treat the prompt text as a discrete parameter $\theta$ of a fixed \emph{DSPy} program and iteratively improve $\theta$ using metric-based evaluation on a training set \cite{khattab2023dspy}. COPRO focuses on instruction-level refinement: starting from a baseline instruction, it generates a pool of candidate rewrites (and, when applicable, output-field prefixes) using a proposer model, executes the student program under each candidate, scores candidates with the task metric, and then updates the instruction by selecting the best-performing candidate before repeating for multiple rounds \cite{dspy_copro_docs}. Operationally, this behaves like a coordinate-ascent style search over prompt text: each iteration proposes localized edits conditioned on prior attempts and their scores, making COPRO a conservative optimizer that tends to steadily improve an initial template without substantially changing the overall program structure \cite{dspy_copro_docs}.

MIPROv2 performs a broader, multi-stage search that can jointly optimize instructions and few-shot demonstrations \cite{dspy_miprov2_docs}. It first bootstraps candidate demonstration sets (e.g., by sampling and/or generating candidate exemplars consistent with the program signature), then proposes a diverse set of candidate instructions intended to capture different task ``dynamics'' (e.g., different ways of specifying constraints, emphasizing format, or prompting reasoning), and finally selects an optimized combination of instruction and demonstrations by maximizing the evaluation metric using Bayesian Optimization over this discrete configuration space \cite{dspy_miprov2_docs}. In practice, MIPROv2 evaluates candidates on minibatches and can use an explicit development split during compilation, which helps scale candidate testing and reduces the risk that a single high-variance batch drives selection \cite{dspy_miprov2_docs}. In our setting, we use MIPROv2 in instruction-optimization mode (optionally with demonstrations), treating the culture-conditioning instruction as the primary tunable component while holding the IVS question block and response-format constraints fixed.

To understand the role of the instruction-proposal model, we run each teleprompter twice: once using a small proposer (Llama~3.2:1B) and once using a large proposer (GPT-OSS:120B). In all cases, the target model $m$ is the model being aligned and is used to generate survey responses for scoring; the proposer model is only used to generate candidate instructions. Figure~\ref{fig:change} visualizes how \emph{DSPy} Culture Prompt Programming (MIPROv2; proposer \texttt{gpt-oss:120b}) changes \texttt{gpt-oss:120b}’s projected position on the Inglehart--Welzel map for each country/territory $c$. Mini-panels (grouped by cultural zone) compare the generic, non-national model projection $\boldsymbol{\mu}_{m,\varnothing}$ (Generic) to the country-aligned projection $\boldsymbol{\mu}^{\mathrm{DSPy}}_{m,c}$ (Aligned), alongside the human reference point $\boldsymbol{\nu}^{\mathrm{IVS}}_{c}$ (Human). The arrow denotes the shift from $\boldsymbol{\mu}_{m,\varnothing}$ to $\boldsymbol{\mu}^{\mathrm{DSPy}}_{m,c}$, while the dashed segment shows the residual discrepancy after alignment. We define the generic and aligned distances to the human benchmark in Equations \ref{eq:nocond_distance} and \ref{eq:dspy_distance}, and summarize the improvement for each $c$ as
\begin{align}
\Delta(c)
&= d_{\text{gen}}(m,c) - d_{\text{align}}(m,c).
\end{align}
Here, $\Delta(c)>0$ indicates that \emph{DSPy} tuning moves the model closer to the human benchmark for $c$. We assess generalization across cultures using 5-fold cross-validation (CV) over countries/territories. In each fold, 80\% of countries form the compilation pool and 20\% are held out for testing. \emph{DSPy} compiler optimizes prompts using the same cultural-distance objective used for evaluation; the held-out country-level split therefore provides the relevant evidence that the best optimized prompts transfer beyond the countries used during compilation. Compilation is performed only on the compilation pool, and for MIPROv2 we further split the compilation pool into training and development subsets for validation during compilation \cite{dspy_miprov2_docs}. We report mean held-out cultural distance:
\begin{equation}
d_{\mathrm{test}}
=
\frac{1}{|\mathcal{C}_{\mathrm{test}}|}
\sum_{c \in \mathcal{C}_{\mathrm{test}}}
d(m,c;\theta^\star).
\label{eq:test_set}
\end{equation}

For Figure~\ref{fig:cultural_distance_dspy} and Table~\ref{tab:fig3_distance_ci}, the DSPy report held-out test-set distances under this CV procedure, whereas the no-culture-prompt and manual culture-prompt rows report the corresponding baseline country-level distances defined in Equations~\ref{eq:nocond_distance}--\ref{eq:manual_distance}.

%% file: sections/results.tex
\begin{figure*}[htb]
  \centering
  \includegraphics[height=0.82\textheight,keepaspectratio]{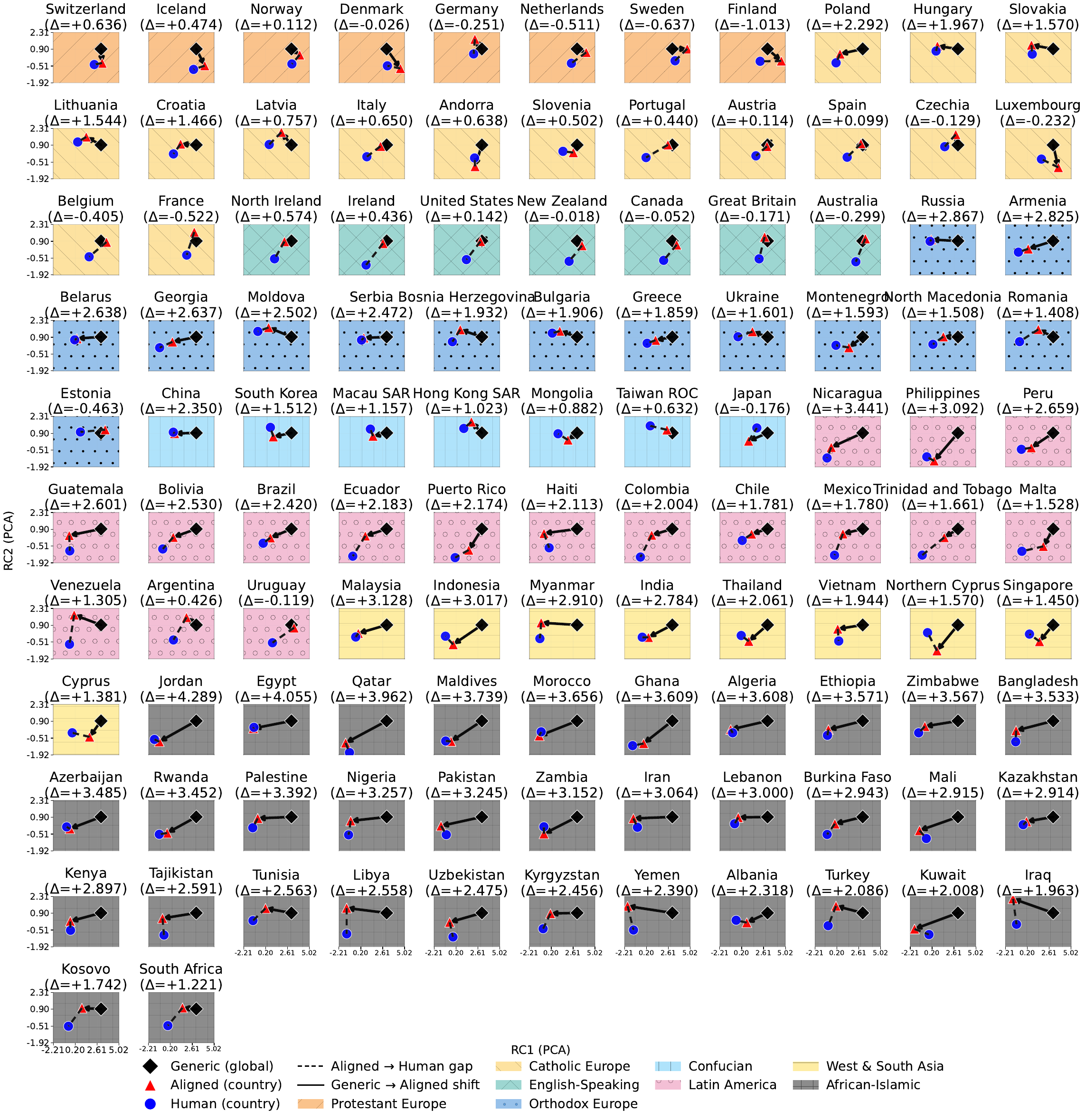}
    \caption{Per-country movement in the Inglehart--Welzel IVS benchmark space (PC1$'$/PC2$'$; Survival vs.\ Self-Expression and Traditional vs.\ Secular) for GPT-OSS:120B after Culture Prompt Programming with \emph{DSPy} (MIPROv2; proposer GPT-OSS:120B). Results are calculated across both compilation and held-out countries over the 5-fold country-level cross-validation runs. Each mini-panel corresponds to one country/territory $c$, grouped by cultural zone using the background tint, and compares the generic non-national projection $\boldsymbol{\mu}_{m,\varnothing}$ (Generic), the aligned country-conditioned projection $\boldsymbol{\mu}^{\mathrm{DSPy}}_{m,c}$ (Aligned), and the human IVS reference point $\boldsymbol{\nu}^{\mathrm{IVS}}_{c}$ (Human). Arrows show the shift from Generic to Aligned, dashed segments show the remaining alignment gap $\lVert \boldsymbol{\mu}^{\mathrm{DSPy}}_{m,c}-\boldsymbol{\nu}^{\mathrm{IVS}}_{c}\rVert_2$, and $\Delta(c)>0$ indicates that alignment moves the model closer to the human benchmark for $c$.}

  \label{fig:change}
\end{figure*}

Figure~\ref{fig:cultural_map_opensource} shows that the generic (non-national) projections of all five open-weight models fall within a relatively tight region of the IVS cultural map, rather than spreading across the full country and territory distribution. This concentration indicates a strong shared default cultural orientation across the models. The model points cluster toward the high self-expression side of the horizontal axis and away from many dense country and territory groupings. This suggests that, when answering IVS items without an explicit national identity, open-weight models tend to draw from a comparatively narrow cultural prior instead of adapting to the broader country-level structure represented in the benchmark. The tight clustering also helps interpret differences among models. Although the model families occupy slightly different positions within the clustered region, these separations are small relative to the overall spread of country and territory locations. This reinforces that the dominant effect under generic prompting is a shared skew, with model-specific differences acting as secondary variation.

This behavior aligns with the proprietary-model results reported by Tao et al.\ \cite{pgae346} using the same cultural-map visualization. In their analysis, GPT-family systems under generic prompting also occupied a compact region of the IVS space that lay nearer to Western cultural clusters than to the global distribution of countries and territories, reflecting a similar default prior in the absence of country conditioning \cite{pgae346}. Our results suggest that this compressed footprint and systematic shift toward the self-expression side is not unique to closed models or a single vendor. At the same time, we observe small but visible separations across open-weight model families within the default region, while Tao et al. emphasized differences primarily across proprietary model versions \cite{pgae346}. Together, these findings suggest that architecture, scale, and training regimes can modulate the precise location of the default prior, but do not remove the overall clustering effect under generic prompting.

\begin{table}[t]
\caption{Mean cultural distance ($\overline{d}$) is reported with 95\% confidence intervals for each model and prompting setting. Distance is measured as Euclidean distance in the IVS cultural-map space between each model projection and the corresponding country/territory reference point $\boldsymbol{\nu}^{\mathrm{IVS}}_c$. For DSPy prompt-programming rows, distances are computed on held-out test-set countries across 5-fold country-level cross-validation and correspond to $d_{\mathrm{test}}$ from Equation~\ref{eq:test_set}. Bold values indicate the lowest distance within each model column, while underlined values indicate the lowest distance within each prompting-setting row.}
\label{tab:fig3_distance_ci}
\resizebox{\columnwidth}{!}{%
\centering
\begin{tabular}{@{}lccccc@{}}
\toprule
\textbf{Prompting setting} & \textbf{llama3.3:70B} & \textbf{llama4:16x17B} & \textbf{gemma3:27b} & \textbf{gpt-oss:20b} & \textbf{gpt-oss:120b} \\
\midrule
No culture prompt & \underline{2.217 ($\pm$ 0.16)} & 2.819 ($\pm$ 0.19) & 3.085 ($\pm$ 0.20) & 2.789 ($\pm$ 0.19) & 3.153 ($\pm$ 0.19) \\
Culture prompt engineering & \textbf{\underline{0.973 ($\pm$ 0.11)}} & 2.006 ($\pm$ 0.12) & 2.008 ($\pm$ 0.13) & 1.688 ($\pm$ 0.11) & 1.778 ($\pm$ 0.10) \\
DSPy MIPROv2, llama3.2:1b & 1.496 ($\pm$ 0.16) & \underline{1.229 ($\pm$ 0.10)} & 2.182 ($\pm$ 0.15) & 1.844 ($\pm$ 0.13) & 1.770 ($\pm$ 0.12) \\
DSPy MIPROv2, gpt-oss:120b & 1.350 ($\pm$ 0.15) & \textbf{\underline{0.818 ($\pm$ 0.08)}} & \textbf{1.499 ($\pm$ 0.13)} & \textbf{1.369 ($\pm$ 0.11)} & \textbf{1.364 ($\pm$ 0.12)} \\
DSPy COPRO, llama3.2:1b & \underline{1.206 ($\pm$ 0.13)} & 1.295 ($\pm$ 0.10) & 2.400 ($\pm$ 0.17) & 1.700 ($\pm$ 0.12) & 2.015 ($\pm$ 0.13) \\
DSPy COPRO, gpt-oss:120b & \underline{1.021 ($\pm$ 0.10)} & 1.163 ($\pm$ 0.09) & 2.250 ($\pm$ 0.16) & 1.695 ($\pm$ 0.11) & 1.860 ($\pm$ 0.13) \\
\bottomrule
\end{tabular}%
}
\end{table}

Figure~\ref{fig:cultural_distance_dspy} reports country-level cultural distance under three prompting regimes: no culture prompt, manual culture prompt engineering, and DSPy-based culture prompt programming. Table~\ref{tab:fig3_distance_ci} provides the corresponding numerical summary by reporting mean cultural distance with 95\% confidence intervals for each model--condition pair; for the DSPy rows, these distances are computed on held-out test-set countries across 5-fold country-level CV and correspond to $d_{\mathrm{test}}$ from Equation~\ref{eq:test_set}. Whereas the figure shows the full distribution of country-level distances using boxplots and overlaid points, the table summarizes each distribution numerically. Consistent with Tao et al.~\cite{pgae346}, distances are largest under generic prompting without culture conditioning, indicating that a single non-national ``default'' response profile remains far from many country/territory benchmarks. This pattern is also visible in Figure~\ref{fig:cultural_map_opensource}, where the generic open-weight model projections cluster in a comparatively narrow, Western-leaning region of the IVS space. Also consistent with Tao et al.~\cite{pgae346}, manual culture prompt engineering reduces distances across all evaluated models: conditioning the persona on the target country shifts the distance distributions downward and narrows dispersion, showing that explicit country identity moves model behavior closer to the intended cultural coordinates, although nontrivial residual variation remains.

Prompt programming with \emph{DSPy} provides a third regime that treats cultural alignment as an optimization objective, but the held-out results show heterogeneous rather than uniformly positive gains. Overall, the strongest results are achieved by MIPROv2 when paired with the larger instruction-proposal model, GPT-OSS:120B. This configuration yields the lowest distance for four of the five target models, while Llama~3.3:70B performs best under manual culture prompt engineering. For Llama~4, all \emph{DSPy} variants improve over manual prompt engineering, suggesting that this model family particularly benefits from replacing a fixed country-conditioning template with a compiled instruction tuned to the cultural-distance metric. For Gemma~3 and the GPT-OSS target models, the gains are more selective: only MIPROv2 with the GPT-OSS:120B proposer consistently outperforms manual prompting, while other \emph{DSPy} configurations provide smaller or negligible improvements.

For country-level observations, we additionally visualize per-country alignment shifts for GPT-OSS~(120B) optimized with MIPROv2 in Figure~\ref{fig:change}. The figure uses the same IVS benchmark coordinate system as Figure~\ref{fig:cultural_map_opensource}, but presents one mini-panel per country/territory, overlaying projections with and without alignment against the human reference point. Figure~\ref{fig:change} should be interpreted as a qualitative diagnostic of how the optimized prompt moves model projections through the IVS space, rather than as the primary estimate of held-out generalization. Unlike the DSPy rows in Figure~\ref{fig:cultural_distance_dspy} and Table~\ref{tab:fig3_distance_ci}, which report held-out test-set distances under 5-fold country-level cross-validation, Figure~\ref{fig:change} pools country-level movement patterns across both compilation and held-out countries. This pooled visualization is useful for interpreting the direction and residual structure of alignment shifts, while the held-out generalization claim is supported by the test-set distances $d_{\mathrm{test}}$ in Table~\ref{tab:fig3_distance_ci}.

A key pattern in Figure~\ref{fig:change} is that the alignment shift is relatively small for several Western countries, such as the United States ($\Delta=+0.142$), but substantially larger for countries whose human benchmarks lie farther from the generic GPT-OSS~(120B) point, such as Jordan ($\Delta=+4.289$). In the IVS coordinate system, Jordan's improvement corresponds primarily to a leftward movement along the Survival vs.\ Self-Expression axis, shifting the model away from its generic high-self-expression position and toward a region of the benchmark space associated with more survival-oriented value profiles. Similar large positive shifts appear for many African--Islamic, West and South Asian, Latin American, and Orthodox European countries, suggesting that prompt programming often reduces distance by moving the model away from its Western-leaning default.

These results imply three broader points. First, the quantitative effects of manual cultural prompting appear to generalize from proprietary systems to open-weight models, reinforcing that generic prompting induces a systematic cultural prior that can be partially corrected by explicit country conditioning \cite{pgae346}. Second, prompt programming can deliver further improvements, but the benefit depends on the optimizer’s search behavior and the capability of the proposal model, with larger proposal models producing more consistently better compiled prompts. Third, the persistence of outliers and the model-dependent nature of \emph{DSPy} gains indicate that cultural alignment via prompting remains incomplete: optimization can improve average alignment while still failing for a subset of countries, highlighting the need for evaluation protocols (such as country-disaggregated cross-validation) that explicitly test robustness and transfer rather than relying on a single global average.

%% file: sections/discussions.tex
Our findings are consistent with several threads in the cultural-awareness literature. First, the persistence of a single Western-skewed ``default'' profile under generic prompting mirrors prior findings on WEIRD-centrism, where model responses tend to align most closely with Western, Educated, Industrialized, Rich, and Democratic populations \cite{atari2023which}. Second, the fact that explicit country identity shifts model behavior toward the target reference location aligns with persona-simulation evidence that nationality conditioning can improve agreement with human responses \cite{kwok2024evaluating}. Third, our observation that alignment gains depend strongly on prompt formulation and optimization setup is consistent with evidence that measured cultural alignment varies with prompt language, persona or demographic conditioning, and evaluation protocol \cite{AlKhamissi2024InvestigatingCA,10852463}. Finally, the model- and country-dependent hard cases that remain after manual prompting and \emph{DSPy} compilation are compatible with the \emph{WorldValuesBench} framing, which shows that performance is not uniform across demographic and cultural conditions \cite{Zhao2024WorldValuesBenchAL}.

This validation-and-extension study has several limitations that motivate follow-on investigation. First, our primary measurement instrument is a small set of forced-choice IVS survey items. This design yields a standardized, population-grounded signal and enables direct comparison to the Inglehart--Welzel benchmark space, but it remains a proxy for cultural alignment rather than a complete measure of cultural understanding. Reducing Euclidean distance in this two-dimensional space does not necessarily imply culturally appropriate behavior in open-ended generation, multi-turn dialogue, document analysis, policy analysis, or decision-support deployments. Thus, downstream document-engineering tasks are needed to test whether these improvements transfer to generated documents and analytic workflows. Second, the gains are heterogeneous: some prompts and optimizers improve average alignment while degrading performance for particular countries or regions. Third, our experiments are English-only and short-form; since results can depend on prompt language and phrasing~\cite{pgae346}, future work should test multilingual, native-language, and longer-form settings. Addressing these limitations would clarify when survey-grounded cultural alignment generalizes beyond questionnaires.

One important direction for downstream evaluation is to test whether survey-grounded cultural alignment improves culturally sensitive document-engineering tasks. In such workflows, LLMs may be asked to summarize policy documents, extract value-laden claims, categorize stakeholder positions, or compare cross-national narratives. Strategic-culture analysis is one example, since it studies how shared beliefs, assumptions, experiences, and patterned behaviors shape perceptions of interests, risks, and acceptable courses of action~\cite{Snyder1977TheSS,johnston1995thinking,Kuznar2023StrategicCulture}, and prior computational work has shown that such patterns can be studied with machine-learning methods~\cite{tappe2021machine}.

%% file: sections/conclusion.tex
We extend survey-grounded cultural alignment evaluation from proprietary systems to open-weight LLMs by reproducing Integrated Values Surveys (IVS)-based projections onto the Inglehart--Welzel cultural map and country-level cultural distance distributions, with and without cultural prompting. Using ten IVS items and benchmark PCA pipeline, we find that open models show systematic cultural skew under generic prompting, and that country-identity prompting reduces misalignment for many countries. We also introduce prompt programming with \emph{DSPy} as an alternative to manual cultural prompt engineering. Prompt programming improves alignment in several held-out settings by reducing Euclidean distance in the two-dimensional IVS/Inglehart--Welzel cultural-map space.